\documentclass[english]{article}
\usepackage[T1]{fontenc}
\usepackage{verbatim}
\usepackage{graphicx}
\usepackage[round]{natbib} 
\usepackage{lscape}
\usepackage{url}
\usepackage{booktabs}

\newcommand{\tableheadline}[1]{\textsc{#1}}

\providecommand{\tabularnewline}{\\}

\makeatother

\usepackage{babel}

\begin{document}

\title{Stereotypical gender actions can be extracted from Web text\footnote{This is a preprint of an
article accepted for publication in Journal of the American Society for Information
Science and Technology copyright \copyright 2011 (American Society for Information Science and Technology).}}

\date{2011-02-26}

\author{Ama\c{c} Herda\u{g}delen\footnote{Current affiliation: New England Complex Systems Institute, Cambridge, USA.}\\
  CIMeC, University of Trento\\
  Corso Bettini 31, Rovereto, Italy\\
  {\tt amac@herdagdelen.com}
  \and
  Marco Baroni\\
  CIMeC, University of Trento\\
  Corso Bettini 31, Rovereto, Italy\\
  {\tt  marco.baroni@unitn.it}}
\maketitle
\begin{abstract}
  We extracted gender-specific actions from text corpora and Twitter,
  and compared them to stereotypical expectations of people. We used
  Open Mind Common Sense (OMCS), a commonsense knowledge repository,
  to focus on actions that are pertinent to common sense and daily
  life of humans. We use the gender information of Twitter users and
  Web-corpus-based pronoun/name gender heuristics to compute the gender bias of the
  actions. With high recall, we obtained a Spearman correlation of 0.47 between corpus-based predictions and a human
  gold standard, and an area under the ROC curve of 0.76 when
  predicting the polarity of the gold standard. We conclude that it is
  feasible to use natural text (and a Twitter-derived corpus in
  particular) in order to augment commonsense repositories with the
  stereotypical gender expectations of actions.  We also present a dataset of 441 commonsense actions with human judges' ratings on whether the action is typically/slightly masculine/feminine (or neutral), and another larger dataset of 21,442 actions automatically rated by the methods we investigate in this study.
\end{abstract}

\section{Introduction}

Online social networks and micro-blogging services are no longer limited to the followers of the latest technologies or teenagers, as might once have been expected. Such technology and services are becoming widely adopted by the mainstream population as an integral part of their daily lives \citep{Fox:2009}. A very prominent example of such an application is Twitter, a micro-blogging service. Twitter lets its users post very short (at most 140-character) messages -- which are called \emph{tweets} -- about what they have been doing or thinking, or what they want to share with their friends and other people. Everyday, tens of millions of tweets are posted by users worldwide. The proliferation of publicly available, user-generated content is a vast source of social data and is already shaping the field of computational social science \citep{Lazer:2009, Thelwall:Buckley:Paltoglou:2010}. 

Another field which enjoys the abundance of Web-based text is knowledge extraction and automated ontology building. An example application is KNEXT (\emph{Kn}owledge \emph{Ex}traction from \emph{T}ext) -- a system proposed for
extracting ``general world knowledge from miscellaneous texts,
including fiction'' \citep{Schubert:Tong:2003}. Web-based text is increasingly used as a source for everyday knowledge (frequently referred as \emph{commonsense knowledge}).  The diversity and sheer amount of text coming from
blogs, Web pages, discussions in newsgroups and other social media
(e.g. Twitter, Facebook) allows one to cover a wide array of domains
and extract a great number of facts.  

A natural way to move forward in this line of research is to augment the Web text with metadata about the users who created them. The highly personal nature of Web-text data, coupled with the rich metadata that are available on sites such as Twitter, promise to add a new dimension to traditional text-based commonsense knowledge extraction. In this study, we are interested in enriching commonsense knowledge with stereotypical expectations of actions.

In a nutshell, our methodology is first to identify verb phrases in a corpus that correspond to commonsense activities, and then extract gender features for a given verb phrase to compute the \emph{gender bias} of the action, which quantifies whether it is a masculine or feminine action. We compare the predicted bias to the stereotypical expectations of
people, collected in a rating task. This allows us to evaluate the performance of our prediction methods.

We focus on two methods to predict the gender bias of a given action. The first one relies on the metadata-guessed gender of Twitter users and their
tweets. In this case, our assumption is that, in Twitter, people talk about the topics and actions that matter to them. In particular, we assume that if an action is frequently mentioned by one gender, this constitutes strong evidence that the action is a typical behavior of that gender. The second method employs a heuristic based on the gendered pronouns and names occurring with the actions observed in a large corpus of Web documents. In this case, the assumption is that if a particular gender is reported to be carrying out an action frequently, then that action is typically associated with that gender.
\section{Background}

\subsection{Common sense}

The holy grail of AI is achieving, if not surpassing, human
intelligence, and there is broad consensus that an explicit
representation of commonsense knowledge will play an essential role in
accomplishing this aim \citep{Lieberman:2008}. Commonsense knowledge has already been exploited in very practical application areas like assisting users of an image retrieval application, by employing knowledge about commonsense activities that relate the terms they search for \citep{Gordon:2001}.

Commonsense knowledge consists of the simple facts that nearly every person knows but almost
never states explicitly because of the very assumption that it is
already shared by everyone. Some examples are that mountains are taller than buildings, grocery has a price, or rivers flow downhill. The assumption that commonsense knowledge is shared is what allows us to
communicate with other people and interact with our surroundings in an
efficient and natural way. Therefore, an AI system needs to possess
common sense if it aims to interact with people in a natural way. In
recognition of this problem, the task of providing the computers with
the most basic but essential bits of everyday knowledge has become a
very hot topic of research in AI and stimulated a number of research
projects such as Cyc and Open Mind Common Sense (OMCS)
\citep{Lenat:1995, Speer:2007}. These projects endeavor to create a
repository of common sense that is represented in a
computationally accessible manner.

Prejudices and stereotypical knowledge present an intriguing aspect of common sense. As human beings, we rely on (and possibly suffer from) stereotypical expectations. Obviously, we would not want to engineer an AI with its own prejudices and stereotypes, but on the other hand, if an AI system is to relate to humans, it should know about the stereotypical expectations as well -- whether it's right or wrong, an AI should know that (we expect that) \emph{women like shopping} and \emph{men like football}. Without an explicit knowledge of the stereotypes, such beliefs can be implicit, hidden and intermixed with other ``objective'' facts in a knowledge base. To quote Sherron (2000) -- who talked about Cyc in particular -- ``Cyc's common sense might very well `believe' certain stereotypical ideas about women, gender, sexual orientation, etc., and then make inferences based on those `beliefs'. Without a strong challenge a homogenizing effect occurs, solidifying the original stereotype among users of the program.'' \citep{Sherron:2000}. We believe our work may help to make such knowledge explicit, thus enabling us to deal with the problem of stereotypical beliefs in commonsense repositories.

\subsection{Corpus-based Gender modeling}

In her influential paper of 1973, Robin Lakoff characterizes some distinguishing aspects of women's speech as being about trivial issues, apologetic, and non-assertive \citep{Lakoff:1973}. In making such claims, her main sources of data are introspection and anecdotal observation. Although her approach has been influential and instrumental in studying the gender differences of language use, later data-driven studies have empirically contested some of her claims \citep{Holmes:1990}. 

More recently, increased access to larger volumes of textual data made it possible to search for (and of course find) more subtle differences in the language use of the two genders. An example of such is the work of \citet{Argamon:etal:2003} which studies the gender differences in formal written texts consisting of samples from the British National Corpus, a collection of British English texts and transcriptions of about 100 million words. Some of their key findings are that females use many more pronouns than males, and males use many more noun modifiers than females.

Another source of textual data for corpus-based studies on gender differences is the Web blogs where people contribute content in a less formal way. It is possible to collect metadata like the gender of the posters, create two sub-corpora based on the text created by the two genders respectively, and then compare the two sub-corpora to reveal statistical differences between the linguistic patterns or content \citep{Argamon:etal:2007}. Such an approach allows us to tap a larger amount of text thanks to the millions blogging about virtually everything on a regular basis. The work of \citet{Argamon:etal:2007} confirms the gender differences previously found in formal texts, and in addition, provides evidence that there are substantial differences in the content as well (e.g., females are blogging about past actions more frequently than males and males are blogging about politics more frequently than females).

Another work on data-driven gender modeling is the study of \citet{Liu:Mihalcea:2007} which is also based on text analysis of blogs. This study is particularly relevant for us because it is based on the gender preferences for several dimensions that are pertinent to user-interface designs (e.g., gender preferences of foods or color and size of things) which happen to be also salient dimensions for commonsense knowledge \citep{Liu:Mihalcea:2007}.

The online social data that is generated by the users of social networking sites have also been used for ``emotion mining'' at large by \citet{Thelwall:Wilkinson:Uppal:2010} who also looked into the age and gender differences in the expression of emotions in public comments of users.

In this study, our approach is similar to the work of \citet{Liu:Mihalcea:2007} in the sense that we are not interested in the gender differences in language use, but rather the differences in the actions that the two genders are associated with. In this case, language acts as a proxy to extract such differences. Compared to their rather exploratory approach, we have a well defined task at our hands -- namely, enriching a commonsense knowledge repository with stereotypical knowledge. Furthermore, blogs posts and Twitter status updates arguably touch different aspects of human life. As we discuss in the next section, the latter is informal, immediate, and entangled with the daily life of people -- making it a more suitable source of commonsense knowledge than blog posts.

\subsection{Twitter and Its Content}
Analyzing the behavior patterns of users in Twitter is hard because they are moving targets. This is partly because people are eager to innovate and
find novel functions for Twitter, or because of company policies of
Twitter -- reflected in the change of the welcoming message to
``Share and discover what's happening right now, anywhere in the world.''
  from ``What are you doing now?''~\citep{Twitter:2009}. Nevertheless, a report published in
2009 by a market research firm, Pear Analytics, provides encouraging
results for our purposes~\citep{Kelly:2009}. The report provides an analysis of the
content of 2000 US-originating tweets in English. The tweets are separated into six categories as news, spam, self-promotion (e.g.
company advertisements), pointless babble (e.g. ``I'm eating a sandwich''),
conversational (e.g., ``hey @joe r u coming 2 dinner 2nite?''), and pass
along value (i.e. retweets). The important result for our purposes is
that the 40.55\% of the tweets are classified as "pointless babble" and
37.55\% as conversational -- which are exactly the type of content we look
for to capture the daily routines of people.
 
In a similar content-analysis study, \citet{Naaman:2010} randomly sampled 3379 personal tweets from the public timeline and classified more than 40\% of the tweets as informational about self at present (e.g. ``i'm tired and upset'', ``just  enjoyed speeding around my lawn on my John Deere. Hehe :)''), and another 10\% as anecdotal about self or others (e.g. ``oh yes, I won an electric steamboat machine  and  a  steam  iron  at  the block party lucky draw this morning!''). These results, again, confirm that people are conveying information about their daily routines in their posts.

These findings motivate our use of Twitter-based text to extract commonsense knowledge because the users of Twitter will frequently talk about actions and concepts that matter to them on a daily basis. As we stated above, common sense is not often explicitly talked about, being part of the background knowledge that enables communication. However, by talking about everyday activities, users are implicitly providing us with fragments of commonsense knowledge (e.g., that you mow lawns).

It is also an interesting task to compare the content of a Twitter-based corpus to a corpus of more traditional Web documents (e.g., ukWaC, to be discussed in subsequent sections). Longer documents such as blog posts, online articles, or documents found on the Web have already been used for commonsense extraction or gender modeling~\citep{Argamon:etal:2003,Liu:Mihalcea:2007,Carlson:etal:2010}. These sources might provide more linguistic contexts about commonsense activities, but on the other hand, such documents will be less directly linked to the events of daily life, and poorer in metadata related to where, when, and by whom they were brought about.
 
\section{Methodology and Data}

\subsection{Overview}
The techniques we propose can work for any kind of phrase, but
at this stage, in order to make a more focused case, we decided to study the actions that play an
important role in our commonsense repository. The commonsense repository OMCS already contains a
substantial number of verb phrases which correspond to actions that humans carry out in a daily
manner. Consequently, we use OMCS as a source of the actions that are
pertinent to the daily life of people. Apart from the semantic
relations between these actions that are provided by OMCS (e.g.,
\emph{motivated by}, \emph{caused by}, \emph{has first subevent}), the stereotypical expectations about the actions are an
important part of our commonsense repository. We propose a methodology
that will allow us to extract \emph{gender features} from large
amounts of text and from the metadata provided by social media (in the
specific case, Twitter).  Subsequently, the gender features are used
to tag the actions in the OMCS repository as \emph{masculine}, \emph{feminine}, or \emph{gender-neutral}.

\subsection{Corpora}
The first corpus we worked on is the Edinburgh Twitter Corpus (ETC) \citep{Petrovic:etal:2010}.
ETC is a corpus of 97 million
tweets, randomly sampled from the Twitter public line during a 4-month
period spanning November 11th 2009 until February 1st 2010. No attempts are made to distinguish or filter according to the language of tweets; therefore it covers a multitude of languages. It contains approximately 1 billion
tokens (i.e., running words, punctuation elements, etc.) contributed by more than 9 million Twitter users. In this study, we limited ourselves to the 34 million tweets that were identified as English, by a heuristic that allows at most 20\% non-English words in a given tweet. The content in Twitter is very colloquial and has some special syntax; therefore, it requires special linguistic pre-processing. We lower-cased and lemmatized all words in Twitter while keeping the emoticons, hash tags, and the user references as is (e.g., the sentence ``The schools were closed.'' became ``the school be close'').

We want to utilize the gender information of the
users, but the corpus does not originally contain the metadata about
the users who posted the tweets. In order to get the metadata, we
queried the Twitter API to obtain the name of
each user. Then, we used the first names of the users to guess
their genders since Twitter does not disclose the gender of the users via its API. We used two lists of the most popular American male and female
names -- compiled from the public data provided by the US Census
Bureau and US Social Security Administration (the lists are made available
at this URL: http://github.com/amacinho/Name-Gender-Guesser). Users whose first names were not on one of the two lists were discarded
from further consideration. Finally, we separated the tweets into two sets according to the \emph{guessed} gender of the users and obtained two sub-corpora containing 82 million tokens for males and 89 million tokens for females (5.2 million male tweets and 5.9 female tweets), each contributed by approximately one million users.

Utilizing the first names of people to guess their gender is inherently a noisy (and
possibly biased) process for several reasons, including bogus
names provided by the users and unisex names. Nonetheless, given the
lack of true gender information, this is the best we can do, and
further sanity checks to be presented in subsequent sections suggest that our guesses make sense in general. Note that gender guessing of names is also a relevant problem in areas like anaphora resolution, and other techniques employed in these domains might be employed in the future, such as name-pronoun co-occurrence, or contextual and morphological cues~\citep{Bergsma:2005,Bergsma:Lin:Goebel:2009}.

We also used ukWaC, a 2-billion-token corpus obtained by a linguistically-informed crawl
of the \texttt{uk} domain conducted between 2005 and 2007, automatically annotated
with part-of-speech and lemma information using the TreeTagger tool
\citep{Baroni:etal:2009}. Compared to ETC, ukWaC is a more traditional
corpus made of relatively long documents, and it does not provide
metadata about who uttered/wrote the collected text and when.
Consequently, we employed linguistic heuristics to extract gender features
as we explain later.

\subsection{Common sense}

The Open Mind Common Sense (http://openmind.media.mit.edu/) (OMCS) project is a commonsense database project initiated by MIT's Media Lab in 1999 \citep{Speer:2007}. It is a collection of several thousands of commonsense assertions. Almost the entire content of OMCS is based on the volunteers' efforts. Since 1999, more than 16,000 people contributed to OMCS via a Web-based interface, resulting in more than 700,000 English facts \citep{Havasi:etal:2009}.

In OMCS, each assertion is represented as a triplet where a relation binds a pair of concepts (e.g., \emph{(apple, AtLocation, apple tree)}). We are not interested in the relations, but only in the concepts. The total number of unique concepts in OMCS is 267,364 -- most of which are multi-word phrases (e.g., \emph{door knob}, \emph{apple tree}). From this larger set, we picked a subset of "actions", i.e., concepts corresponding to
verb phrases. In this context, a verb phrase is simply a
multi-word expression beginning with a verb. In turn, a verb is
any word that is tagged much more frequently as a verb than any
other part-of-speech in the ukWaC corpus. The number of actions obtained from OMCS this way is 49,754. As we will observe, OMCS also contains some spurious or meaningless concepts, like ``do what'' in the assertion \emph{(preserve, AtLocation, do what)}. We do not attempt to filter these concepts/actions.

\subsection{Gold standard}

As our gold standard dataset, we randomly sampled 702 phrases from the
set of actions detected in the ETC Twitter corpus and represented in OMCS. We give the details of phrase
detection in the next subsection. We used
Crowdflower's (http://www.crowdflower.com) crowd-sourcing services to
have the verb phrases annotated by people. We used a 5-point scale in
the annotation task: Typically feminine (-2), slightly feminine (-1),
neutral (0), slightly masculine (1), and typically masculine (2), with
an extra option of ``Not a verb phrase''/``meaningless''. Each rater
was presented a verb phrase and was asked to provide his/her opinion
about the phrase. 
After the data
collection phase, we eliminated 97 phrases which were rated by less than five raters. Among the remaining 605 phrases, the average agreement probability of a rater with the majority answer for a phrase was 44.29\% -- after excluding the rater's own response. For a 6-way decision task, the random baseline would have an expected value of 16.67\%. For subsequent analysis, we further discarded another 164 phrases whose majority designations were ``meaningless''. This
filtering left us with 441 phrases. Overall, 112 raters contributed (each rater annotated at least 18 verb phrases). We did not keep track of the gender of the raters.

For each verb phrase in our dataset, we
calculated the \textit{gender score} as the mean score of the
responses it received. This score serves as the \emph{human} gold
standard for the stereotypical gender expectation of the corresponding
action. For illustrative purposes, we provide a stratified random
sample of the gold standard dataset in
Table~1.


\begin{table}[th]
\centering
\begin{tabular}{rcrc}
\toprule
\tableheadline{Feminine} & \tableheadline{Mean score} & \tableheadline{Masculine} & \tableheadline{Mean score}\tabularnewline
\cmidrule(r){1-2}\cmidrule(r){3-4}

become nurse& -2.00  &ask stupid question & 0.20\tabularnewline
make doll & -2.00 & join circus & 0.33\tabularnewline
freshen up & -1.60 &generate revenue & 0.71\tabularnewline
get assistance & -1.33 & impress people & 1.00\tabularnewline
feel cold & -1.20 & try solve problem & 1.00\tabularnewline
choose love & -0.80 & enjoy power & 1.29\tabularnewline
pick berry & -0.60 & see pretty girl & 1.60\tabularnewline
put weight & -0.40 & catch football & 2.00\tabularnewline
resolve problem & 0.00 & want woman & 2.00\tabularnewline
\bottomrule
\end{tabular}
\caption{A stratified random sample of the gold standard.}
\label{tab:stratified-sample}
\end{table}

\section{Corpus analysis methodology}
\label{sec:methodology}
\label{ssec:phrase-detection}

We searched for the OMCS phrases (both actions and non-actions) in the lemmatized version of
the corpora, allowing at most
one intermittent token between two lemmas of the phrase (e.g., the
concept \emph{long holiday} is said to be \emph{observed} when we
encounter the text ``long holidays'' or ``longing for holiday'').

In the case of ETC, in order to compute the gender bias of an action, we detect all of its corresponding verb phrase's utterances in the male and female sub-corpora and compute the proportion of the number of male utterances as the raw scores. We report normalized scores computed using expectation and variance values from a binomial distribution with $p$ equal to the overall proportion of male utterances and $n$ equal to the total number of utterances for the action at hand. Formally, if we respectively denote the number of male and female occurrences of a particular verb phrase with $m$ and $f$ (subject to $m+f=n$) and the total number of male and female occurrences of all verb phrases with $M$ and $F$ (i.e., $M=\sum{m}$ and $F=\sum{f}$) then the final formula to compute the normalized score $s$ of the given verb phrase becomes $s=(m-p)/\sigma_{m}$ where, $p=M/N$ and $\sigma_{m}=\sqrt{np(1-p)}$. The rationale of using the proportions of gender utterances is that in Twitter people talk about what they do; hence, if a certain verb phrase is used, proportionally, more often by one gender then we conclude that it is probably a more typical action of that gender -- we discuss this point later in this section. As a convention, we arbitrarily picked the sign of the measure so that a masculine (feminine) bias results in a positive (negative) score. A gender bias of 0 means the two genders are equally likely to mention a given verb phrase. 

The ukWaC corpus does not contain information about the gender of the
people who use a given phrase. Instead, we employ a heuristic that uses the gender information of the pronouns and the proper names. Whenever we detect a verb phrase in a sentence, we look for the nearest pronoun or proper name (identified by the part of speech) in the sentence to the left
side of the phrase and if it is a ``he'' or a male name (``she'' or a female name) we count the
occurrence of the phrase as a male (female) utterance -- the gender of the names are guessed by using the same lists we employed for Twitter users. For example the sentence ``She became a nurse.'' would be counted as a female utterance for \emph{become nurse} whereas ``Jack tried to solve the problem.'' would be counted as a male utterance for \emph{solve problem}. The gender bias is then computed by using the proportions of male and female utterances, in a similar fashion as for the Twitter corpus. 

We should note that the two approaches are quite different in their nature. The Twitter approach directly taps the gender information about who uses a given verb phrase -- regardless of who actually carries out the corresponding action. An initial analysis revealed that half of the pronoun-associated verb-phrase utterances in Twitter are used with the first person singular pronoun ``I'' -- indicating that people often talk about what they do, but also about what other people do. Whether people do (or like to do) what they talk about frequently is an open empirical question that we will partially address here. On the other hand, the ukWaC approach allows us to tap a larger
amount of text and follows a more anecdotal path. It does not try to
guess the gender of who utters a phrase, but the gender of the person
that is said to perform the action described by the phrase. 

\section{Results}
\subsection{Sanity Checking}
An important limitation of the current methodology is that we guess the gender of the users by their first names. Unfortunately, the lack of a gold standard of the genders of users also avoids a thorough evaluation of the accuracy of the guessing method. Nevertheless, it is possible to carry out some sanity checks to make sure the results are in accordance with what we expect. In Table~2, we can see the distribution of male and female users who mention the phrases \textit{my husband}, \textit{my boyfriend}, \textit{my wife}, and \textit{my girlfriend} at least once in their tweets. The phrases relating to the male significant others or partners are mentioned much more frequently by females and vice versa. Obviously, this is not a direct proof but evidence suggesting the gender guessing method is reasonably accurate.


\begin{table}[htbp]
\centering
\begin{tabular}{lll}
\toprule
\tableheadline{Phrase} & \tableheadline{Males (\%)} & \tableheadline{Females (\%)} \\
\midrule
my husband & 7 & 93 \\
my boyfriend & 10 & 90 \\
\midrule
my wife & 87 & 13 \\
my girlfriend & 73 & 27 \\
\bottomrule
\end{tabular}
\caption{The frequency of male and female users who mention specific phrases (based on unique user count). The frequencies are computed on a balanced sample of Twitter users that have equal number of male and female users.}
\label{tbl:my-husband-my-wife}
\end{table}

\subsection{Commonsense coverage}

Although our focus is primarily on the actions contained in OMCS, it is informative to look at the coverage of the common sense in our corpora as a whole. The percentage of the commonsense concepts that are represented in OMCS and observed in corpora is quite high: Out of the original 267,364 OMCS phrases (including actions and non-actions), we observed 54\% (145,486)
in ukWaC and 51\% (136,128) in ETC. If we focus on the actions then out of the 49,754 verb phrases, we observe 47\% (23,455) in ukWaC, and 43\% (21,442) in ETC, associated with a gender bias. Although the filtered ETC is much smaller than ukWaC, its coverage is about the same.

\subsection{Gender bias}

The top 10 masculine, feminine, and neutral actions computed over the two corpora are given in Table~3. The results in the table look indeed encouraging for the effectiveness of a corpus-based approach. Interestingly, there is no overlap between the two corpora -- suggesting that Twitter and ukWaC may be covering different aspects of common sense. As we remarked earlier, OMCS contains spurious or meaningless phrases like ``do in'' or ``do so''. Especially in ukWaC, such phrases that include ''do`` seem to have a very strong masculine gender bias, presumably because of the frequent use of ``he'' either as a masculine or as a gender-neutral pronoun used to refer to people in general in the same underspecified contexts in which do is used. 

In the next three subsections, we will evaluate the gender bias in detail. First, we look at
the correlation between the corpus-based predictions of gender score
and the human gold standard. In the second subsection, we analyze the predictive
power of the gender bias on the direction of the stereotypical gender
expectations of actions. After that, we carry out qualitative
analyses that help us to interpret the data.
In all cases, the Twitter and ukWaC scores are converted to z scores.

\begin{landscape}
\begin{table}[thp]
 \tabcolsep 4.8pt
\begin{tabular}{lllllll}
\toprule
 & \multicolumn{2}{c}{\textsc{Masculine}} & \multicolumn{2}{c}{\textsc{Feminine}} & \multicolumn{2}{c}{\textsc{Neutral}}\tabularnewline
\cmidrule(r){2-3}\cmidrule(r){4-5}\cmidrule(r){6-7}
Rank & Twitter & ukWaC & Twitter & ukWaC & Twitter & ukWaC\tabularnewline
\midrule
1. & make money & do so & go bed & give birth & buy cheese  & want much\tabularnewline
2. & get free & take over & feel like & take place & chew food & ask yourself why\tabularnewline
3. & want make money & think himself & want go & find out & go sea & build nuclear weapon\tabularnewline
4. & make playoff & do in & feel good & become pregnant & wait area & build product\tabularnewline
5. & unite state & do to & make smile & let know & take seat & carry everything\tabularnewline
6. & earn money & keep himself & make cry & see herself & wait show & enter exit\tabularnewline
7. & operate system & do what & go school & take part & lose track time & establish priority\tabularnewline
8. & go down & become king & go sleep & provide information & make mind & prepare depart\tabularnewline
9. & try out & go close & come home & think herself & become true & sell magazine\tabularnewline
10. & come soon & raise up & see new & add basket & know appreciate & teach read write\tabularnewline
\bottomrule
\end{tabular}
\caption{Top ranking gendered actions collected from Twitter and ukWaC. The neutral actions have the smallest absolute gender bias. Note that the two corpora do not have a single verb phrase in common in the top ranking lists for either gender.}
\label{tbl:top-ranking-gendered-actions}
\end{table}
\end{landscape}
\subsection{Spearman correlation}


In Table~4, we report the Spearman correlations between the gold standard and the
gender biases computed by various methods. For each row, the number
in parenthesis is the number of items covered by the corresponding
method. \emph{Combined} refers to taking the average of Twitter and
ukWaC scores for each verb phrase. In this method, only the items that
are covered both by Twitter and ukWaC are used. \emph{Matching signs} reports the performance of combined method only on the verb phrases for which Twitter and ukWaC scores agreed on the sign of the gender biases.


%
\begin{table}[th]
\centering
\begin{tabular}{llll|}
\toprule
\textsc{Method} & \textsc{Spearman} & \textsc{Coverage}\tabularnewline
\midrule
ukWaC & 0.27 & 98\% (433)\\
Twitter & 0.28 & 100\% (441)\\
Combined & 0.33 & 98\% (433)\\
Matching signs & 0.47 & 52\% (231)\\
\bottomrule
\end{tabular}
\caption{Spearman correlations between various corpus-based scores and the gold standard. Coverage is the percentage of items that have an associated gender bias for the corresponding method. All reported correlations are statistically significant for $p<0.001$}.
\label{tbl:spearman}
\end{table}

\subsection{Predictive power of the gender biases}

To further assess the predictive power of the corpus-based scores, we
transformed our gold standard into a two-class dataset by using the sign of the gender scores of the verb phrases as their labels -- discarding the verb phrases with a human score of zero. Thus, \emph{gold-standard} phrases were labeled as \emph{masculine} and \emph{feminine}, respectively. We used the sign of the gender biases based on Twitter and ukWaC as the predicted labels, and computed the area under the ROC curve (AUC) and accuracy of the predictions. The AUC of a classifier is the area under the curve defined by the corresponding
true positive rate and false positive rate values obtained by varying
the threshold of the classifier to accept an instance as positive.
Intuitively, AUC is the probability that a randomly picked positive
instance's estimated posterior probability is higher than a randomly
picked negative instance's estimated posterior probability
\citep{Fawcett:2006}. In our case, the masculine and feminine labels are not inherently ``positive'' and ``negative''. To train a binary classifier, one should arbitrarily designate one of the labels as positive and the other as negative. The results do not depend on which one is picked as positive. We report both accuracy and AUC because while accuracy is easily interpretable (i.e., the percentage of verb phrases that were correctly labeled), its precise value is dependent on the threshold chosen for the classification of the instances. AUC provides a more robust way of characterizing the performance of the classifier without having to pick a particular threshold.

\begin{table}[th]
\centering
\begin{tabular}{llll}
\toprule
\textsc{Method} & \textsc{AUC} & \textsc{Accuracy} & \textsc{Coverage} \%\tabularnewline
\midrule
ukWaC & 0.64 & 0.61 & 99\% (375)\tabularnewline
Twitter & 0.65 & 0.61 & 100\% (380)\tabularnewline
Combined & 0.67 & 0.59 & 99\% (375)\tabularnewline
Matching signs & 0.76 & 0.70 & 54\% (205)\tabularnewline
\bottomrule
\end{tabular}
\caption{Classification performance of various corpus-based scores. Random baseline for both measures is 0.50. Coverage is the percentage of items that have an associated gender bias for the corresponding method.}
\label{tbl:auc}
\end{table}


The results tabulated in Table~5 show us that it is possible to separate the feminine and masculine actions from each other with a reasonably high success. Especially if we limit ourselves to the phrases for which
both Twitter and ukWaC scores agree on their signs, we can achieve an
AUC of 0.76 and an accuracy of 0.70. 

\subsection{Qualitative pattern analysis}

So far, we have seen that the corpus-based gender biases are not
perfect predictors of the human gold standard, but are still reasonably valuable indicators of stereotypical expectations. We were
able to obtain a Spearman correlation of 0.46 and an AUC of 0.76 on
a restricted subset of the dataset (more than half of the OMCS
sample). Considering that we have more than 20,000 verb phrases in the
general dataset sampled from OMCS, we can expect to gender-tag
approximately 10,000 actions reliably -- assuming that for half of the phrases the two methods will agree on the sign, and that the performance we observe on our sample is representative.

Another important point is that our errors are instructive. In
Figure~1, we see the scatter plot of
Twitter versus ukWaC scores of the verb phrases and we observe
that the two methods have quite different results. The Spearman correlation between the scoring methods is only 0.19.


%
\begin{figure*}[!htb]
\includegraphics[width=1\columnwidth]{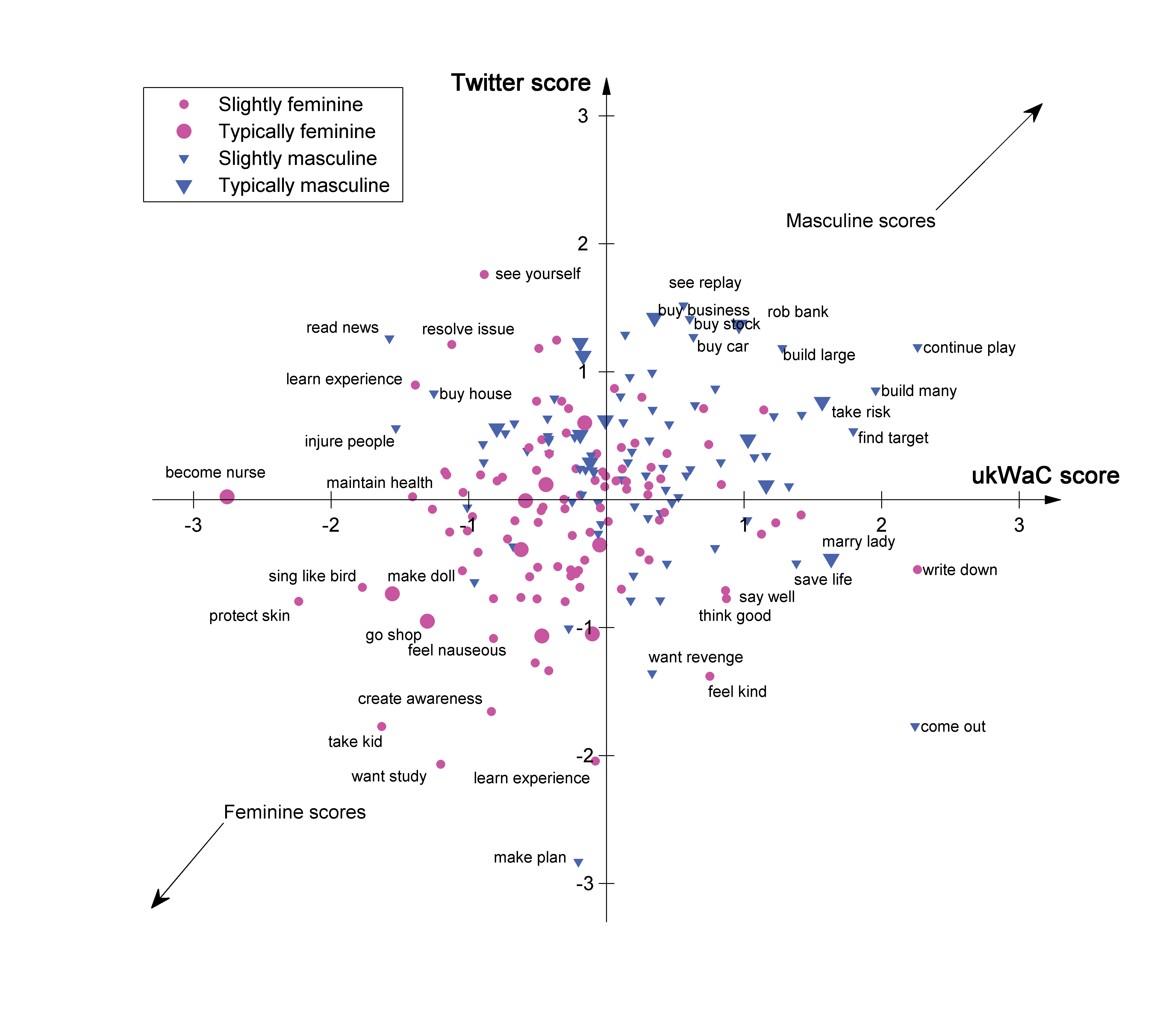}\caption{ukWaC score (x-axis) versus Twitter
score (y-axis); for both axes more positive values represent more masculine gender bias. Color coding represents the polarity of the human gold standard and size is proportional to the magnitude of average human score.}
\label{fig:ukwac-twitter}
\end{figure*}

The first and third quadrants of the figure, where the signs of Twitter and ukWaC biases match, contain a majority of gold-standard masculine and feminine actions, respectively. This is not surprising as we already saw the increased performance of the matching-signs method. It is more interesting to look at the mismatch between Twitter and ukWaC. The actions that are placed in the second and fourth quadrants have mismatching signs for the Twitter and ukWaC biases. A plausible interpretation for such items is that they are the actions that one gender talks about much, but is reported to be doing less often. Consider ``resolve issue'' in the fourth quadrant for instance -- an action that is rated as slightly feminine by the human raters: in ukWaC, females are reported to be ``resolving issues'' more often than males (hence a negative ukWaC bias), whereas in Twitter, males mention this action more frequently (hence a positive Twitter bias). A similar case is ``come out'', an action that is rated as slightly masculine by the human raters, which is placed in the second quadrant: in ukWaC, males are reported to be ``coming out'' more often, whereas it is female Twitter users who mention this action more frequently. 

Another informative visualization is given in Figure~2 which is a scatter plot of the human gold
standard versus Twitter bias. In this figure, we plot only the actions with a sign mismatch. We can interpret these actions as the actions that one gender talks about a lot (mentions it in Twitter more frequently than the other gender does) but are rated to be associated
with the other gender, by the human raters. For example, the actions \emph{take note} and \emph{get assistance} are mentioned more often
by males in Twitter but they are considered to be feminine by the human raters. The phrases \emph{build snowman} and \emph{want revenge} are examples of the gold-standard masculine actions that females talk about.

%
\begin{figure*}[!htb]
\includegraphics[width=1\columnwidth]{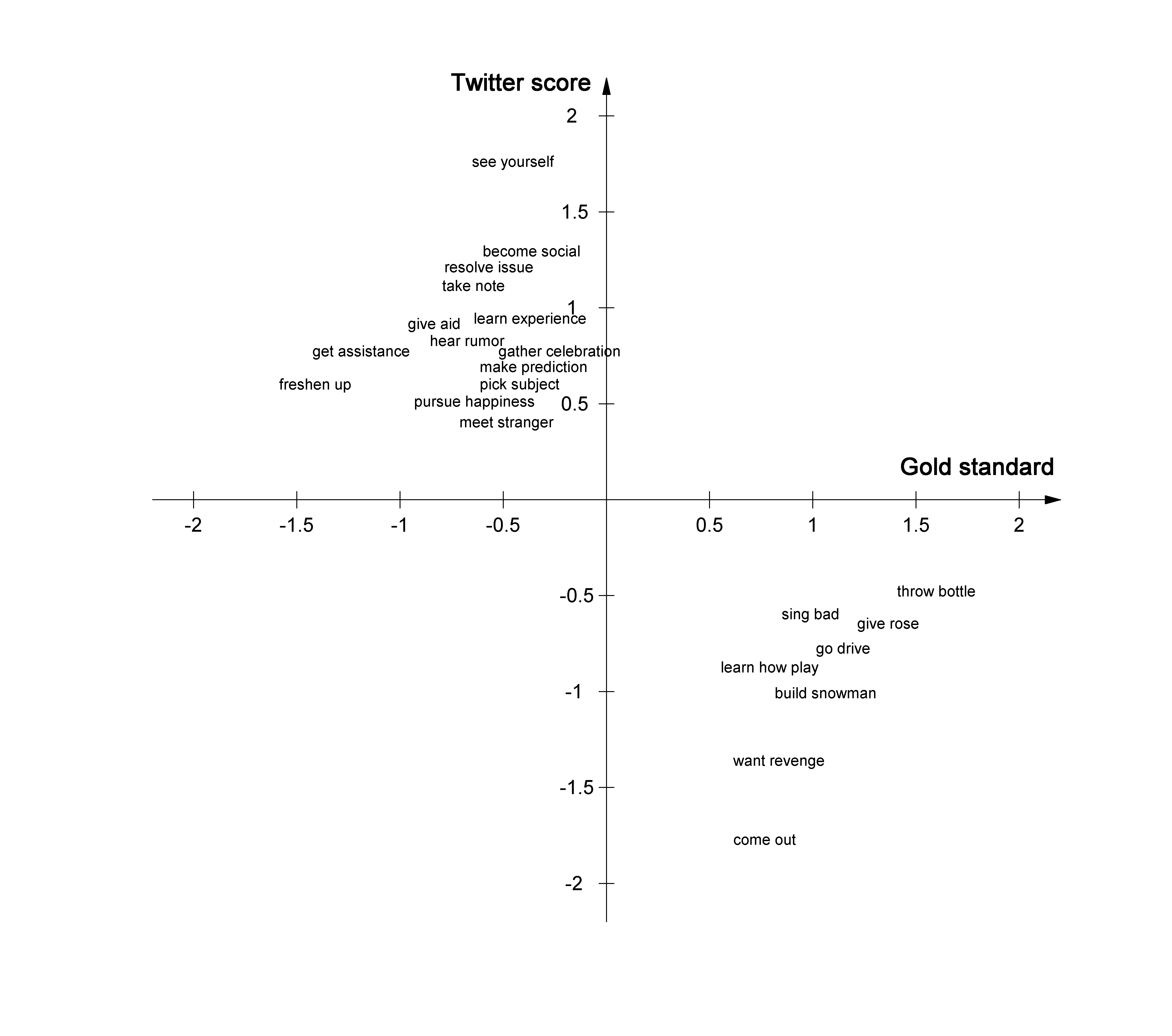}%
\caption{\label{fig:say-dont} Twitter bias (y-axis) versus the gold standard (x-axis). Only the actions that are in the second and fourth quadrants are shown.}
\end{figure*}

In sum, the qualitative analysis suggests a more complex picture, in
which mismatches (between Twitter and ukWaC, as well as corpora and
the human gold standard) are not necessarily mistakes of the
text-based methods, but a sign that we are tapping into different
kinds of information: explicit assessments of stereotypical actions
(gold standard), actions that males and females are reported as being
doing in natural written discourse (ukWaC), and actions that they like
to talk about (Twitter). Each of these sources (and their combinations) might be useful for
different purposes.

\section{Conclusion}

We introduced novel ways to utilize the metadata
contained in a Twitter corpus and simple linguistic cues in the ukWaC
Web corpus in order to extract stereotypical expectations about
actions that are pertinent to common sense. While working on this
problem, we observed that both ukWaC and Twitter have a wide coverage
of commonsense concepts. More than half of the unique concepts
in OMCS were detected in the two corpora. The gender-filtered Twitter
corpus is much smaller than ukWaC, but has an equally wide coverage of
common sense.

In this study, we focused on a small sample of the actions that are represented in OMCS. However,
the corpus-based methods we presented compute the gender biases of all of the actions that are represented in
OMCS. The dataset containing all 21,442 actions extracted from OMCS with the Twitter and ukWaC scores, and the human ratings we used as gold standard for the smaller sample can be downloaded from this URL: https://github.com/amacinho/Gender-Expectations-Predictions/

We conclude that both the metadata about the Twitter users and corpus-based gender attribution heuristics definitely help in stereotypical knowledge mining. On several performance metrics, the
Twitter-based approach is at least as good as the ukWaC-based scoring system. Moreover,
combining the two methods works even better.

Apart from predicting the gender expectation of a given verb phrase (which
seems feasible, based on the experiments we reported), the methodology
may allow us to dig in deeper and provide refined data that might be
used in sociolinguistics, gender studies, and personalized information retrieval and recommendations. The demographic
dimensions we can extract from Twitter (or other similar social media) are not limited to gender. Time
of the day, geographical location, and other metadata can be employed
in similar ways to augment commonsense knowledge repositories, and
equip computers with an even better understanding of how humans work.

\bibliographystyle{abbrvnat}
\bibliography{main}

\end{document}